# Evaluation of a Bi-Directional Methodology for Automated Assessment of Compliance to Continuous Application of Clinical Guidelines, in the Type 2 Diabetes-Management Domain

**Authors:**

Avner Hatsek

Department of Software and Information Systems Engineering, Ben Gurion University, Beer Sheva, Israel

Declarations of interest: none

Irit Hochberg

Endocrinology, Diabetes, and Metabolism Institute, Rambam Health Care Campus, Haifa, Israel

Bruce Rappaport Faculty of Medicine, Technion - Israel Institute of Technology, Haifa, Israel

Declarations of interest: none

Deeb Daoud Naccache

Endocrinology, Diabetes, and Metabolism Institute, Rambam Health Care Campus, Haifa, Israel

Declarations of interest: none

Aya Biderman

Department of Family Medicine and the Siaal Research Center for Family Medicine and Primary Care, Faculty of Health Sciences, Ben-Gurion University of the Negev and Clalit Health Care Services, Southern district, Beer Sheva, Israel

Declarations of interest: none

Yuval Shahar

Department of Software and Information Systems Engineering, Ben Gurion University, Beer Sheva, Israel

Declarations of interest: none

# Abstract

Evidence-based recommendations are often published in the form of clinical guidelines and protocols, as documents intended to be used by clinicians to provide the state of the art care. However, as demonstrated repeatedly in multiple clinical domains, clinicians often do not sufficiently adhere to the guidelines in a manner sensitive to the context of each patient. Such gaps are important to detect; fast, large-scale detection might lead to specific adjustments, usually of the clinicians' management patterns, but possibly of the guidelines themselves.

In this study, we evaluated the *DiscovErr* system, in which we had implemented a new methodology for assessment of compliance to continuous implementation of clinical guidelines. This new methodology is based on a bi-directional search from the objective of the guideline to the longitudinal multivariate patient data, and vice versa. The evaluation of *DiscovErr* was performed in the type 2 Diabetes management domain, by comparing its performance to a panel of three clinicians, two experts in diabetes-patient management and a senior family practitioner highly experienced in diabetes treatment. The system and the three experts commented on the management of 10 patients who were randomly selected before the evaluation from a database containing longitudinal records of 2,000 type 2 diabetes patients. On average, each patient record spanned 5.23 years; the overall data of the selected patients included 1,584 time-oriented medical transactions (laboratory tests or medication administrations). We assessed the correctness (i.e. precision) and completeness (i.e. recall or coverage) of the comments provided by the *DiscovErr* system relative to the gold-standard comments of the clinicians. After providing their own comments regarding the therapy provided by the original physician treating the patient (typically, a general practitioner), the clinical experts assessed both the correctness and the importance of each of the *DiscovErr* system comments. The completeness of the system was computed by comparing its comments to the ones made by each of the experts or by their majority. The *DiscovErr* system provided 279 comments. The three experts made a total of 181 different unique comments. The completeness of the *DiscovErr* system was 91% when the gold standard was comments made by at least two of the three experts, and 98% when compared to comments made by all three experts. A total of 172 comments were evaluated by the experts for correctness and importance: all of the 114 medication-therapy-related comments, and 35% of the 165 tests-and-monitoring-related comments (by selecting at random three of the patient records). The correctness of the system was 81% when compared to comments that were judged as correct by both of the diabetes experts, and 91% when compared to comments judged as correct by one diabetes expert and at least as partially correct by the other. Regarding the importance of the system comments, 89% of the comments were judged as important (versus less important) by both experts, 8% were judged as important by one of the two experts, and 3% were judged as less important by both experts.

When we computed completeness and correctness scores based on all the comments made by the human experts, augmented by system comments that were validated, the completeness scores of the experts were 75%, 60%, and 55%; the expert correctness scores were respectively 99%, 91%, and 88%. Thus, if it were considered as an additional, fourth expert, the *DiscovErr* system would have ranked first in completeness and second in correctness (when computing these measures also for the human experts, using a majority consensus of opinions as the gold standard).

We conclude that systems such as *DiscovErr* can be effectively used to provide medical critique and assess the quality of continuous guideline-based care of large numbers of patients.

**Highlights**

- Evaluation of a new approach for automated guideline-based quality assessment
- Evaluated on 1584 records of 10 randomly selected type 2 diabetes patients
- The system's critique was compared to that of three diabetes experts
- The automated system demonstrated 91% to 98% completeness, and 91% correctness
- The system ranked 1st in completeness and 2nd in correctness, compared to the experts

# 1. Introduction

## 1.1. The Necessity of Assessing the Quality of Medical Care

The necessity of common standards for medical care is becoming increasingly clear to the medical community. Evidence-based recommendations are published world-wide in the form of clinical guidelines and protocols. These guidelines are usually published in a text format, and are intended to be used by clinicians to provide the state of the art care. Evidence also indicates that implementation of these guidelines may improve medical care and reduces its costs [Grimshaw and Russel 1993; Micieli et al., 2002; Quaglini et al., 2004; Patkar et al., 2006; Ruben et al., 2009].

However, as demonstrated repeatedly in multiple clinical domains, clinicians and patients often do not sufficiently and uniformly adhere to the clinical guidelines in a manner that is sensitive to the context of each patient. A typical recent example is the type 2 diabetes domain, in which significant variance has been demonstrated across five European countries with respect to the application of the American Diabetes Association (ADA) and the Kidney Disease: Improving Global Outcomes (KDIGO) guidelines with respect to metabolic and blood pressure control, and the use of renin–angiotensin system–blocking agents, statins, and acetylsalicylic acid [Eder et al., 2019]. Another example is the Preeclampsia-Toxemia (PET) domain, in which compliance of clinicians, when not assisted by a guideline-based decision-support system (DSS), had a completeness, out of the total guideline-based recommendations relevant to the patient at hand, of only 41% to 49%, rising to 93% when guideline-based recommendations were first suggested by the DSS; furthermore, 68% of the clinician proposed actions were correct, but redundant, when compared with the patient-specific recommendations of the guidelines and the patient record; the redundancy was reduced to 3% when the clinicians were first exposed to the DSS suggestions [Shalom et al., 2015].

Such gaps in adherence to clinical guidelines are important to detect; efficient, automated, large-scale detection might lead to fast, specific, focused adjustments, usually of the clinician management patterns, but possibly of the guidelines themselves.

Efforts had been made to develop automated systems that can perform some form of evidence-based assessment of the quality of care, the major examples include the HyperCritic [van der Lei, Musen 1990], Trauma-AID [Clarke et al., 1993], Trauma-TIQ [Gertner, 1997], AsthmaCritic [Kuilboer et al., 2003], IGR [Boldo, 2007], RoMA [Panzarasa et al., 2007], and the model checking for critiquing [Groot et al., 2008].

In addition, a growing number of medical centers and organizations have established internal quality and risk assessment units that perform quality assessment of the medical treatment, typically on a random subset of the patient population, or in dire circumstances in which mistakes have already been made. These risk assessment units usually examine the medical records manually, or by using relatively simple computational methods, and compare the medical records to a deterministic set of quality measures that are created specifically for that purpose.

## 1.2. The Objective of this Research: Evaluation of the *DiscovErr* System

The main objective of the current study, is to perform a quantitative realistic evaluation of the *DiscovErr* system, which we had previously designed and implemented [Hatsek and Shahar, 2021; Hatsek 2014]. The *DiscovErr* system implements a new approach for automated guideline-based quality assessment of the care process, a methodology for bidirectional knowledge-based assessment of compliance to an evidence-based guideline. The methodology assesses the degree of compliance (using a form of fuzzy temporal logic for computing the degree of partial matching) when applying clinical guidelines, with respect to multiple different aspects of the guideline (e.g., the guideline's process and outcome objectives). The assessment is performed through a highly detailed, automated quality-assessment retrospective analysis, which receives as input a formal representation of the guideline and of the multivariate longitudinal patient data, and returns as output a series of critiquing comments that represent gaps between the intended guideline and the actual longitudinal record. These comments (see example in figure 1) point out missing actions, redundant actions, incorrectly selected actions, duplicate actions, actions performed too early or too late, etc.

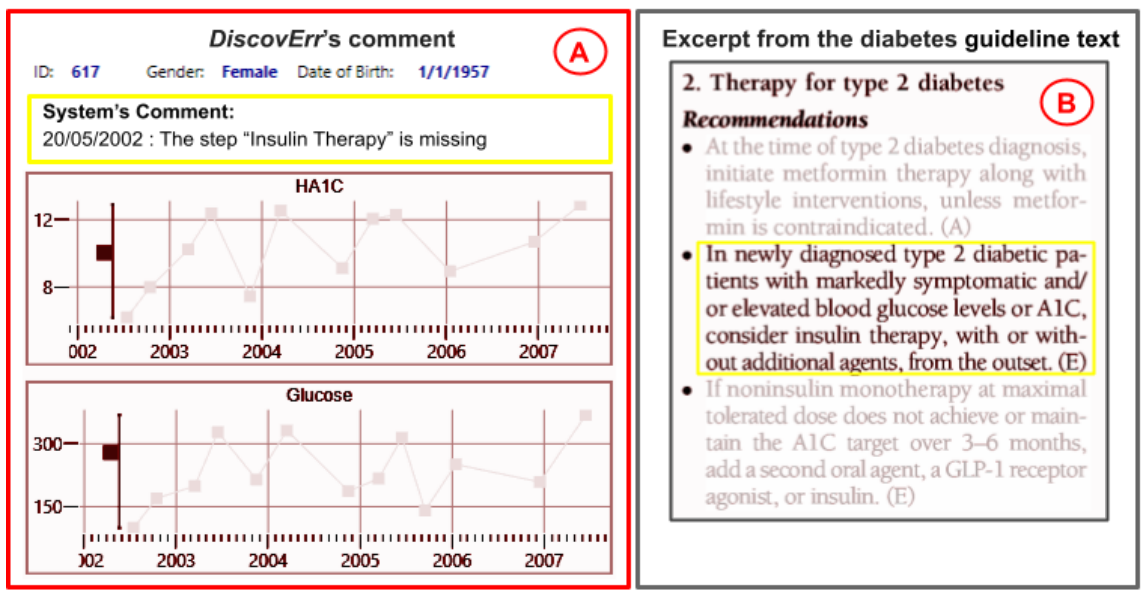

**Figure 1.** An example of a comment made by the *DiscovErr* system (A), regarding missing insulin therapy for a newly diagnosed patient with very high levels of HbA1c (>10%) and Glucose (>290 mg/dL), and the relevant excerpt from the original guideline (B).

For the formal representation of the guideline, the *DiscovErr* system uses the Asbru language [Miksch et al., 1997; Shahar et al., 1998], due to its formal representation of multiple types of process and outcome intentions as temporal patterns to be *achieved, avoided*, or *maintained*, and of multiple types of explicit *conditions*, such as several types of eligibility conditions and of guideline-completion conditions. The Asbru language had been successfully used in several projects for formal representation of clinical guidelines in multiple clinical domains, some examples include: Neonatal Intensive Care [Fuchsberger et al., 2005], Hyperbilirubinemia in healthy term newborns [Sips et al., 2006], breast cancer [Groot et al., 2008], preeclampsia / toxemia of pregnancy [Shalom et al., 2015], gestational diabetes, and atrial fibrillation [Peleg et al., 2017]. The temporal semantics of Asbru have also been shown to be consistent and clear [Schmitt et al., 2006].

Our evaluation methodology for assessing the *DiscovErr* system is most similar in its spirit to the well-known evaluation of the HyperCritic system [van der Lei et al., 1991], a system that critiqued the care of patients who had hypertension. HyperCritic was evaluated with the help of a panel of eight expert physicians, whose critiques of the care provided by general practitioners were compared to those of the system. However, the computational methodology underlying the *DiscovErr* system is quite different from that underlying the HyperCritic system, and is based directly on a formal representation of the original clinical guideline.

## 2. Methods

The evaluation of the *DiscovErr* system was designed to assess the feasibility of implementing it in a real clinical quality-assessment setting. The experiment was performed in the diabetes domain, in which we specified knowledge from a state-of-the-art guideline in a formal representation, and applied the system to real retrospective patient data, to assess to what extent the treatment complied with the standard guideline. The data were presented to expert physicians. The expert physicians were then asked to manually evaluate the compliance of the treatment of a set of patients to the guideline by examining the electronic medical records. Then, the comments given by the system regarding compliance to the guidelines were presented to the experts, and they assessed the correctness and importance of these comments.

The general objective of the evaluation was to enable assessing the correctness (i.e., precision) and completeness (i.e., recall or coverage) of the comments provided by the system relative to the gold standard comments of the clinicians, when automatically analyzing electronic medical records for guideline-based compliance of the therapy manifested by these records.

Note that the terms completeness and correctness used here refer to different measures than the formal definitions of first-order logic, and describe a continuous grade for the level of coverage and level of correctness of the comments given by the system regarding guidelines compliance issues.

### 2.1. Research Questions

The general objective of the research was to evaluate the feasibility of applying the system in realistic clinical settings. This led to the definition of several more specific research questions, that aim to assess the quality and significance of the results of the compliance analysis. For each research question presented in this section (i.e., specific objective), we describe here the general idea of how it was measured, as more details are available in the experimental design and the results sections.

**Question 1: Completeness: Does the system produce all or most of the important comments relevant to the task of assessing compliance to a guideline?**

To use such a system in real clinical settings, it is necessary to evaluate the quality of its results. One dimension of the quality of the results is the completeness, or level of coverage, as it is important to know if all or most of the deviations from the guideline are detected by the system.

To answer this question, we conducted an experiment in which the system and three medical experts examined the medical records of the same set of patients, and provided comments regarding the compliance to the same clinical guideline for the management of diabetes mellitus. The measure for completeness was defined as the portion of compliance-related comments that were mentioned by the system from the comments mentioned by the majority of the medical experts.

**Question 2: Correctness: Is the system correct in its comments regarding the compliance to the guideline?**

Another dimension of the quality of the results is the level of correctness of the system output, sometimes referred to as precision. It is important to know the proportion of correct compliance comments provided by the system out of all the system comments.

**Question 3: Importance: Are the comments provided by the system significant and important for understanding the quality of treatment?**

In addition to measuring the correctness and completeness of the compliance analysis results, we were also interested in measuring the level of significance of the comments provided by the system regarding the compliance.

To answer questions 2 and 3, we conducted an experiment in which two diabetes experts evaluated the correctness and importance of the compliance-related comments provided by the system when analyzing the medical records of a set of patients. The measure for correctness was defined as the portion of system comments that were evaluated as correct by the two experts. The measure for importance was defined as the portion of system comments that were evaluated as important by the two experts.

**Secondary research issues**

In addition to the primary research questions presented above, we were also interested in examining two secondary issues that focus on the human aspects of the experiment, and to answer them to some extent.

**Issue 1:** Comparison of the quality-assessment comments of the experts and the system: What is the completeness and correctness of the comments of the expert physicians, when using the comments of the colleagues of each of the experts as a gold standard? How does the performance of the *DiscovErr* system measure up, using this method?

**Issue 2:** Similarity in the experts' meta-critiquing: What is the level of agreement between several experts regarding the quality of the system assessments?

We found these issues interesting, as one of the major motivations for clinical guideline implementation, is its contribution to the reduction of the variance among treatments provided by different physicians. It was important for us to learn whether the experts mostly agree with each other when performing the task of compliance assessment, which is a different task from providing a real treatment. Due to the fact that in the compliance analysis task, the experts refer to the same guideline, we assumed that the experts would have a reasonable level of consensus.

In addition, evaluating the level of agreement between the experts is also essential in order to enable the evaluation of the system itself, as there is no true meaning for the experts' evaluation of the system if there is no agreement between them.

To address the first issue, we defined indirect measures to evaluate the completeness and correctness of the experts themselves. The specific details of these measures are described in the Results section. To address the second issue, we used Cohen's Kappa statistic, and applied it to the evaluations of the experts regarding the correctness and importance of the system comments (i.e., as part of addressing Research Questions 2 and 3).

## 2.2. Experimental Design

To answer these research questions, we designed a study that included several experimental steps (see Figure 2), which enabled evaluation of several aspects of the overall framework.

A preliminary step before designing the study was to find a medical domain. After examining several medical domains, we decided to perform the experiment in the Diabetes domain. This domain is suitable for clinical guideline application, with well-established evidence, and in which the diagnostic and therapeutic process is performed over a sufficiently long period of time (i.e., a period of time of sufficient duration for performing meaningful longitudinal patient monitoring and therapeutic decisions that might adhere to a continuous-care guideline). It is important to mention that although we wished to evaluate the system in more than one medical domain, we had limited resources and decided, instead of performing several smaller experiments in several medical domains (collaborating with one expert in each domain), to focus more deeply on a larger experiment within a single domain.

We obtained a dataset that included anonymized information about 2,038 patients diagnosed with type 2 diabetes. The data were obtained as part of a research collaboration with the Quality Assessment Department of a large Israeli HMO, and included 378,273 time-oriented data records, including details about test results and medication orders and purchases that are relevant for assessment of compliance to the diabetes-management guideline. The records in this data set covered up to five years of continuous treatment for each of the patients, together with general demographic information about the patient, such as gender and age.

In the following sections, we describe the structure of the study and each experimental step.

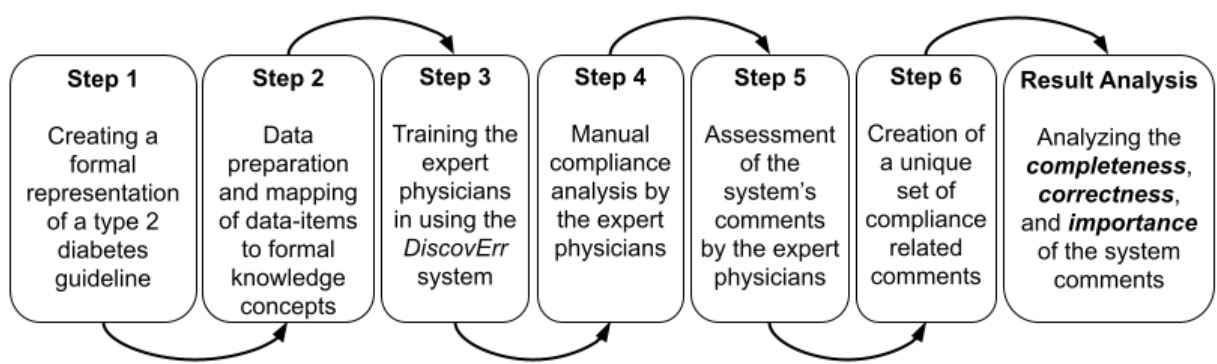

**Figure 2.** A diagram of the experimental steps in our evaluation of the *DiscovErr* system.

## Step 1: Creating a formal representation of an established diabetes-management guideline

The first step in the evaluation focused on the formal representation of a clinical guideline within the selected clinical domain, using the knowledge specification interface of *DiscovErr*, which uses, for *procedural knowledge* representation, such as recommendations regarding the performance of hemoglobin A1c monitoring under certain conditions, the Asbru language; and for *declarative knowledge* representation, such as for definition of different abstractions of the Blood-Glucose-value raw-data concept into abstract values such as Hypoglycemia, Normoglycemia, or Hyperglycemia, the k*nowledge-based temporal-abstraction* (KBTA) ontology [Shahar 1997]. The guideline that was selected was the *Standards of Medical Care in Diabetes* [American Diabetes Association, 2014], that roughly corresponded to the time in which the patients were managed. This is a comprehensive guideline that is based on an extensive literature review and is updated every year. The guideline addresses multiple aspects of diabetes, from screening through diagnosis to multiple aspects of the long-term treatment.

The overall guideline specification was performed in two steps. In the first step, we created a prototype version of the formal guideline by detecting the relevant sections in the guideline, deciding on the optimal representation according to the formal model, and using the system knowledge acquisition interface to formally specify the knowledge. In the second step, one of the expert physicians was involved, and assisted in validating the knowledge represented in the first prototype and extending it with additional knowledge that was not explicit in the original guideline. An example of such knowledge that did not exist in the original guideline is the definition of the concept "reduced kidney function", which is important for the management of Metformin drug therapy. The definition of this concept was not mentioned in the guideline and had to be added by the expert by referring to additional sources. Another example of a decision that was made by the expert physician during the specification process, is a general decision regarding the specification of the temporal aspect of recommendations about patient monitoring. In the first prototype, when specifying a recommendation about the temporal aspect of periodic monitoring (e.g., test hemoglobin A1c every 3 months), we used the temporal annotations of the Asbru language, "earliest-start" and "latest-start", to define a flexible time range for the next measurement. After consulting with the expert physician, the "earliest-start" was removed in some sections of the guideline to prevent the system from providing comments regarding tests that were performed too early. This was decided because in some cases in clinical practice, such as hemoglobin A1c and LDL monitoring, there is no limit to the number of tests (frequency of testing) of the same clinical variable that the physician can order. This relaxed constraint is of course true only for certain tests, which are simple to perform, are not too expensive, and are useful for assessing response to a treatment change and improvement of patient engagement.

The process of formal representation of the guideline was concluded after applying the system to a small set of longitudinal patient records, and briefly examining the results to perform the validation (note that since the objective of the system is not to manage patients, but rather to perform a retrospective assessment of compliance, we did not need to extend the performance evaluation beyond a reasonable assurance that the representation seems accurate).

The overall knowledge specification process was completed in two weeks, in which the first week was dedicated for the first step of the process, of creating a prototype representation, and the second week was dedicated for collaborating with the medical expert to validate and improve the guideline's representation.

**Step 2: Preparation of the data and mapping it to concepts in the formal knowledge base**

The data preparation is a preliminary step that enables application of the system on the medical records. To allow the system to retrieve the relevant items in the patient data at run time, there is a need to map between the concepts in the guideline's formal knowledge base and the terms in the patient database. This can be completed only after finalizing the specification of the guideline in a formal representation, when the complete set of relevant concepts is determined.

Regarding tests and measurements, the original data set did not include codes from standardized vocabularies for medical concept identification. Therefore, we had to manually examine each term in the database, and map it to the relevant declarative concept in the system knowledge-base, and if necessary, perform a conversion to the same units of measurement specified in the knowledge case. Examples of relevant test and measurement concepts include blood glucose, hemoglobin A1c, blood cholesterol, and creatinine lab tests.

Regarding drug therapy concepts, the data included codes according to the standard of WHO's ATC classification system, a fact that simplified the mapping process. Using the interface of *DiscovErr*, we only had to select the relevant ATC terms and attach them to the concepts appearing within the appropriate steps in the formal clinical steps' library. For example, the concept used within the step "initiate-insulin-therapy" was mapped to several potentially relevant ATC items, such as "A10B: Insulins and analogues for injection, fast-acting" or "A10AC: Insulins and analogues for injection, intermediate-acting". Mapping the concepts appearing within the clinical steps to these higher level classes of the ATC hierarchy allows the system to automatically detect the drug identifiers found in the electronic medical record and to relate them to the relevant items in the knowledge.

**Step 3: Training the expert physicians**

A training session was conducted with each of the expert physicians. The session covered the following topics:

- A general overview of the research and its specific goals.
- Presentation of the diabetes guideline, which was provided to the experts in a printed copy format and in an electronic format; in both formats, the relevant sections were visually marked. It is important to note that all of the experts were familiar with the selected guideline, so this topic was covered within a short time.
- Description of the patients' dataset, its source, structure, available clinical variables, and data format.
- Demonstration of the *DiscovErr* system and its comment-making interface, by performing a compliance analysis of two or three demonstration patients. We found that the demonstration allowed the experts to better understand the idea of the system, and to understand the nature of the comments provided by the system regarding compliance to the guideline.
- A short training regarding the specialized user interface that we had created solely for this evaluation, which was used by the expert physicians to both insert their own evaluations of the actions found within the patient record, and to add their evaluation of the assessment performed by the system of these actions (see the interface description below).

**Step 4: Manual compliance analysis of the patients' management, by the expert physicians, performed on a randomly selected set of patients**

Following the training step, the experts were provided with a convenient (visual) interactive interface for browsing the complete set of longitudinal data of multiple types of a randomly selected subset of patients from the database that was introduced earlier to them, for the full duration of time for which the patient was followed (up to five years per patient). They were then asked to perform the two evaluation tasks (i.e., (1) assessment of the quality of care given by each patient's care provider, and (2) assessment of the quality of the *DiscovErr* system comments).

Specifically, we first asked each of the expert physicians to review the data of the same 10 randomly selected patients, which comprised, altogether, 1584 time-oriented records, and manually add comments regarding the compliance of the patient treatment to the diabetes guideline. At this point, the experts could not yet see the system comments regarding the compliance of these patients to the guideline. They were provided with a user interface for the visualization of the raw temporal data of the patient, and for adding their comments using this interface.

For each of their comments, the experts were asked to provide data about the date of the clinical event, the importance/significance of the clinical issue, the type of the comment, and a textual description of the clinical issue. For the type of the comment they could select from a given set of comment types, or insert their own type of comment. The given set of values is presented in Table 1.

The experts were not limited regarding the time they could invest for the evaluation of each patient, except their own limitation on time constraints; they could browse several years of data for each patient, and add as many comments as they wanted.

**Step 5: Assessment, by the expert physicians, of the comments provided by the system**

In this step, we asked the two diabetes experts to evaluate the comments of the system regarding the compliance to the diabetes guideline. The experts used the output interface of the system, to explore the guideline compliance comments. The comments were presented together with the relevant patient data in a visual manner. Using the meta-commentary part of the user interface, the experts assessed the comments given by the system regarding correctness and importance (see Figure 3). For each system comment, the expert marked whether the comment is *correct* according to the guideline when considering the data that was available to the care provider at the specific point in time. In addition, the expert marked whether the comment is *important*, to express an opinion regarding the level of clinical significance of the issue referred to by this comment (regardless of whether the comment itself is judged as correct or incorrect). For the correctness evaluation, the expert could select from three options: correct, partially correct, and not correct; for the importance evaluation, the expert could select from two options: important and less important. (The binary scale emerged from the discussions with the experts; the pilot experiment included originally an additional level, very-important, but it turned out to create problems in maintaining consistency when assessing the system comments: for example, different instances of the same comment regarding late HA1C lab tests were annotated both as important and as very-important by the same expert). In addition to annotating regarding the correctness and importance of the system comment, the experts could enter an optional free text comment when they wanted to add additional information.

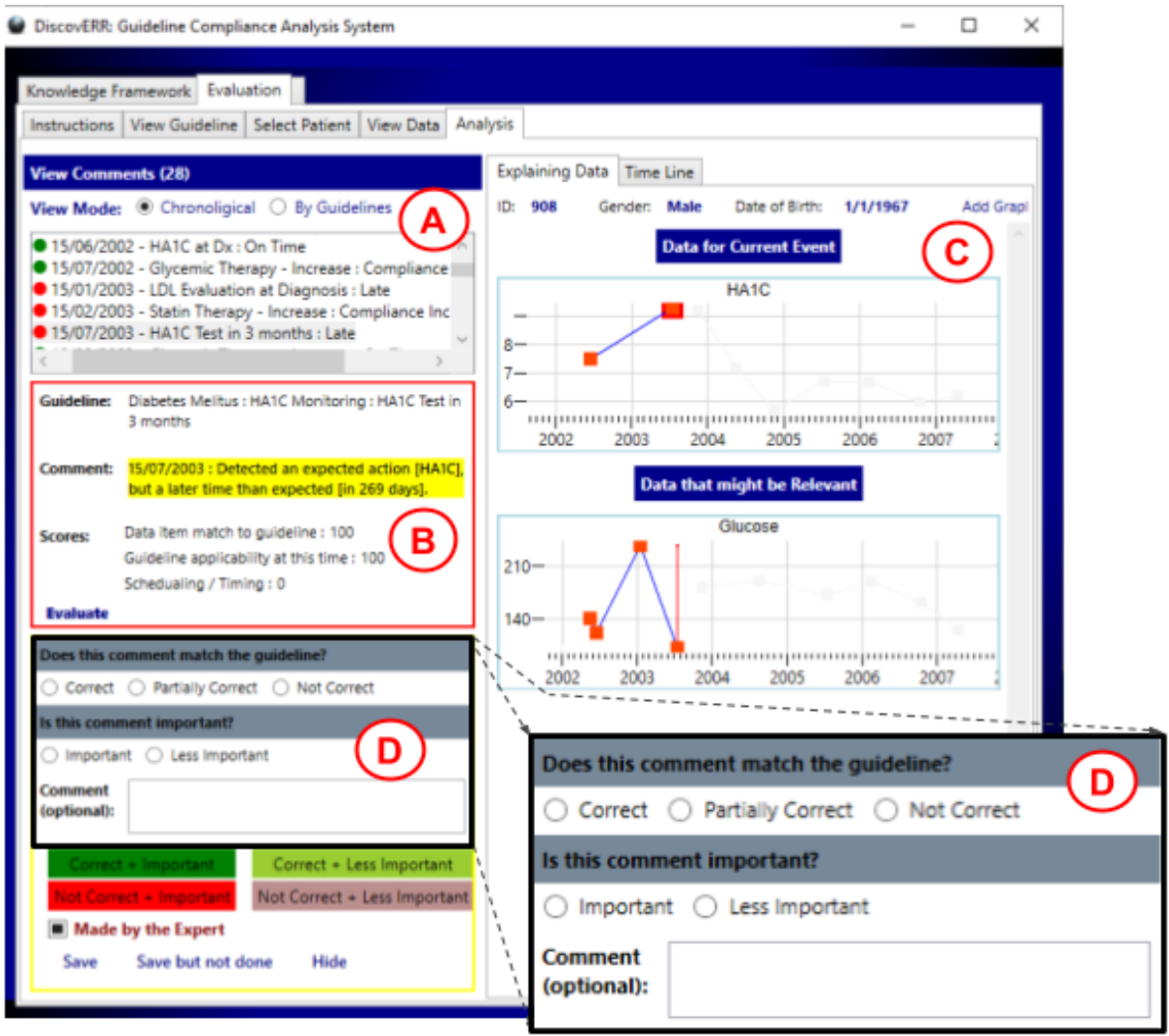

**Figure 3.** The interface used by the experts for evaluating the comments of the system. The expert can view the list of comments on the top left side of the screen (A). When selecting a specific comment, additional explanation is displayed (B) together with graphs showing raw data (C). The explanation includes the specific path of the guideline and a textual description with details about the system scores. The graphs are of one or more parameters that lead the system to its comment. Zoom into the part of the UI for meta-critiquing (4), shows the manner in which the experts expressed their opinion of the *DiscovErr* system comments.

**Step 6: Creation of a unique set of compliance-related comments**

The final step of the result evaluation included additional preparation that was required for supporting a deeper quantitative analysis of the results. In this step, the knowledge engineer scanned and analyzed each of the comments resulting from the previous steps of the evaluation, both of the system and of the experts, and extended it with additional metadata.

**Cross-tabulation and annotation of the experts' comments**

In the first step of this preparatory analysis, we examined the comments that were added by the experts in the step of manual evaluation of the management of the treatment of patients. We examined each comment given by each expert, and compared it to the comments of the other two experts and to those of the system. Each comment was then annotated with three additional attributes that indicated whether it was detected by (1) the system, (2) the other first expert, (3) the other second expert. For example, when examining a comment of the first diabetes expert, for each patient we scanned the full set of comments of the system, of the second diabetes expert, and of the family physician expert; and annotated whether the comment was included in each of these comment sets. By opening four instances of the system, we scanned hundreds of comments provided by each expert, and annotated whether each comment was detected by the system and by the other experts.

It is important to mention that in this part of the preparatory analysis, we had to deal with cases in which comments were phrased differently by each expert. This was solved by manually reviewing the comments and relating to the semantic meaning of each comment. For example, the same scenario of a (too) late clinical action can be described as a "late action" by one expert, or as a "missing action" in the earlier period of the patients record by the other expert, based on the same data and the same guideline-based clinical knowledge. Thus, the result of this phase was essentially a determination of a set of (semantically) unique comments.

An additional issue that had to be addressed in the preparatory analysis of the expert comments is the fact that some of the comments provided by the experts were regarding knowledge that belonged to sections of the source guideline that were not included in the formal representation, or that belonged to other clinical guidelines. Comments of this type were annotated as "out-of-scope" and were excluded from the formal analysis of system completeness.

**Cross-tabulation and annotation of the evaluations given by the experts on the system comments**

In the second step of the preparatory analysis, we re-examined the comments provided by the system. As explained earlier the system comments were already evaluated in the previous step regarding correctness and importance by each of the diabetes experts. However, to support additional levels of result analysis, we re-examined each comment to determine whether it exists in the sets of patient management comments provided by each diabetes expert when they made their own management compliance comments, before they were exposed to the comments of the system. In this manner, each comment of the system was extended with additional meta-attributes that annotated whether it was, in fact, mentioned by each of the two diabetes experts.

# 3. Results

Recall that our main objectives in the current study were to assess the *completeness* of the *DiscovErr* system comments, relative to the comments that need to be made; the *correctness* of the comments made by the system; and the *importance* of the comments that were made or missed by the system. We also wanted to examine several secondary issues, such as the completeness and correctness of the medical experts themselves, and their internal agreement .

## 3.1. Completeness of the System Comments

The results in this section relate to the first part of the experiment, namely, the manual compliance analysis of the patient management, by the three expert physicians, performed on a randomly selected set of patients. The results in this section provide insights about several interesting aspects; these aspects include the time needed for the experts to complete the compliance analysis task, the distribution of the types of compliance issues found by the experts in the records of the patient management, a comparison between the experts based on their compliance comments, the completeness of the comments given by the system relative to one or more comments given by the experts, and more. Our objective in this phase of the evaluation was to measure the *completeness* of the *DiscovErr* system comments, namely, the portion, out of the comments mentioned by the experts, that were made by the system (we shall present a more precise definition later).

On this first part of the experiment, which included two diabetes experts and one family medicine expert, each expert evaluated the longitudinal record of each patient within a group of 10 patients who were randomly selected before the evaluation. The average time period of patient data was 5.23 years. The overall data of the 10 patients consisted of **1,584** time-oriented single medical transactions (e.g., a single laboratory test result or a single administration of a medication), i.e., an average of 158 time-oriented medical transactions per patient, which had to be examined by the experts. The mean time for an expert to examine a single patient was 27 minutes. The experts made a total of 381 comments, from which 31 were labelled as insights (i.e., comments that are not directly related to guideline compliance), 21 were out of the scope of the guideline's sections that were used in the experiment, and 329 were compliance comments within the scope of the guideline. A total of 66% of the comments were regarding drug actions and 34% regarding test and monitoring actions.

Table 1 presents the distribution of the comments given by the experts regarding the type of compliance comments. Note that "Action is on time" is the only positive type comment, denoting the fact that in the expert's opinion, the action is a correct one and was performed in a timely manner.

Table 1. The distribution of the comments given by the experts with respect to the type of compliance issue.

| Compliance Issue Type | Comments | % |
|---|---|---|
| Late Action (Expected action that was performed too late) | 118 | 36% |
| Action is On Time (An action expected by the guideline is performed on time) | 61 | 19% |
| Missing Actions (Missing expected action) | 59 | 18% |
| Patient Compliance ((Low compliance to medications) | 56 | 17% |
| No Support (Action should not be started at this time) | 32 | 10% |
| Redundant (Another action with same intention performed previously) | 1 | 0.3% |
| Early Action (Action performed too early) | 1 | 0.3% |
| Guideline Contradicted (Action that contradicts the guideline's recommendation) | 1 | 0.3% |
| All | 329 | 100% |

To analyze the completeness of the system, the comments were divided into three groups according to the level of support by the experts: comments mentioned by only one expert, by exactly two experts, or by all three experts; then, by meticulously examining the text of all comments, the number of unique compliance issues was counted for each group, where a unique issue is a specific clinical management topic that can be mentioned, possibly using other words, or as part of another comment by one or more experts and the system. There were 50 (27.5%) unique compliance issues that were mentioned by all three; 48 (26.7%) of the unique compliance issues that were mentioned by two experts; and 83 (46%) of the unique compliance issues that were mentioned by only one expert. In total, 98 (55%) of the 181 unique compliance issues were mentioned by two or more experts out of the three, i.e., by a majority of the experts.

Table 2 displays the completeness results analyzed for each level of support.

**The completeness, as defined in the context of our evaluation, was expressed by the portion of unique compliance issues detected by the system, relative to the overall number of unique compliance issues mentioned by a group of experts.**

Applying in each case a proportion test, the completeness by the system of the unique compliance issues mentioned by all three experts was found to be significantly higher than the completeness by the system of the unique compliance issues that were mentioned by only two experts (z = 2.51, p = .012), which, in turn, was found to be significantly higher than the completeness of the system with respect to unique compliance issues that were mentioned by only one expert (z = 2.11, p = .035).

Table 2. Completeness of the comments given by the experts relative to the unique compliance issues, by level of support of the three experts.

|  | Unique Compliance Issues | Compliance Issues Detected by System | Completeness Regarding the exact no. of supporting experts | Completeness at that support level or higher |
|---|---|---|---|---|
| Mentioned by one Expert | 83 | 55 | 66% | 80% |
| Mentioned by two Experts | 48 | 40 | 83% | **91%** |
| Mentioned by three Experts | 50 | 49 | 98% | 98% |
| All | 181 | 144 |  |  |

The system completeness for comments that had the support of a majority of the experts (i.e., two or three out of three) was 91%, and the completeness was significantly higher for comments that had higher levels of support by the experts.

### 3.2. Correctness of the System Comments

The results in this section relate to the second part of the experiment, in which the two diabetes experts evaluated the correctness and importance of the compliance comments mentioned by the system. This evaluation was performed on the data of the same 10 patients who were examined by the experts in the first part of the experiment. The mean time for the experts to evaluate the comments provided by the system on a single patient was 9.5 minutes.

In this phase of the evaluation, we wanted to determine the **correctness of the *DiscovErr* system comments, defined as the proportion of correct comments provided by the system, out of all of the comments made by the system.**

The system provided a total of 279 comments when applied to the data of all 10 patients, 165 (59%) regarding issues related to the compliance to tests and monitoring recommendations, and 114 (41%) regarding issues related to the compliance to medication therapy recommendations. A total of 172 comments were evaluated (62% of the comments by the system): 100% of the 114 medication-therapy-related comments and 35% of the 165 tests-and-monitoring-related comments. The reason for this discrepancy is that only the data of three of the patients were fully evaluated regarding the correctness of the tests and monitoring comments made by the system, due to constraints on the experts' time.

To validate the consistency of the expert assessments of the comments given by the system, we started by calculating the level of agreement about the validity of these comments between the experts themselves, before proceeding further in the analysis. We measured the inter-expert agreement using a weighted version of Cohen's Kappa coefficient, assigning linear disagreement weights according to the distance between the ordered values of the correctness ordinal scale (correct, partially-correct, not-correct). Cases in which both experts agreed were assigned a weight of 0; cases in which the experts did not agree, but with a distance of only one level between the assessments (i.e., correct/partially-correct, partially-correct/not correct), were assigned a weight of 1; and cases in which the experts did not agree, with a distance of two levels between the assessments (correct/not-correct), were assigned a weight of 2.

The weighted Kappa was 0.61, a value that represents a good agreement [Altman 1991], and is significantly higher compared to a chance value of 0 ($p < .05$).

In addition to the Kappa, we measured the level of agreement between the diabetes experts regarding the truth value of the correctness of the system comment that was being assessed (whether agreeing that the system is correct or agreeing that the system comment is incorrect). The experts fully agreed on the correctness, partial correctness, or incorrectness of 151 of the 172 evaluated system comments (87.8%), partially agreed on the truth value of an additional 20 of the 172 comments (11.6%) (i.e., the comment was evaluated as partially correct by one expert and as correct or not-correct by the other expert), and did not agree at all on only one of the system comments (0.6%).

Since it was clear that the experts significantly agree on the correctness or incorrectness of the system comments, we now looked at what they actually said about these comments. According to the judgment of diabetes expert 1, 84% of the comments were correct, 11% were partially correct, and 5% were not correct; according to the judgment of diabetes expert 2, 88% of the comments were correct, 7% were partially correct, and 5% were not correct. The next step was to integrate the results of the two diabetes experts.

Table 3 displays the correctness when integrating the result of both diabetes experts. To integrate the evaluation of the two experts, the possible combinations of the correctness results were organized into six combination groups. The groups were ordered from top to bottom, from the most correct combination to the most incorrect combination. For each combination the number of comments is presented together its proportion to the total number of comments.

To present the complete picture, Table 3 shows the full results for all levels of agreement of the experts with the comments made by the *DiscovErr* system. The cumulative percentage represents the proportion between the number of comments that are at least as correct as the current combination to the total number of comments; thus, 81% of the evaluated comments were judged as correct by both experts, while 91% of the evaluated comments were judged as correct by both experts or judged as correct by one expert and partially correct by the other. Only 3% of the evaluated comments were judged as not correct by both experts.

Thus, 91% of the comments given by the system were fully supported by at least one of the experts, while the other expert did not disagree with the system (i.e., partially or completely agreed with its comments). We consider that 91% portion as representing, for practical purposes, the level of correctness of the system comments.

Table 3. Correctness of the system comments according to both diabetes experts.

|  | Comments | % | Cumulative % |
|---|---|---|---|
| Comments judged as correct by both experts | 139 | 81% | 81% |
| Comments judged as correct by one expert and as partially correct by the other | 17 | 10% | 91% |
| Comments judged as partially correct by both experts | 6 | 3% | 94% |
| Comments judged as not correct by one expert and as correct by the other | 1 | 1% | 95% |
| Comments judged as not correct by one expert and as partially correct by the other | 3 | 2% | 97% |
| Comments judged as not correct by both experts | 6 | 3% | 100% |
| All | 172 | 100% |  |

### 3.3. Importance of the System Comments

Regarding the importance of the guideline compliance issues that were referred by the comments given by the system; 153 comments (89%) were judged as referring to important issues by both experts, 14 comments (8%) were judged as referring to important issues by only one of the experts, and 5 comments (3%) were judged as less important by both experts.

We measured the inter-expert agreement regarding the importance assessment, using the standard (0/1 weights) version of Cohen's Kappa coefficient, and the Kappa coefficient was 0.37. This value, although significantly higher than a chance value of 0 ($p < .05$), technically represents only a fair agreement [Altman, 1991], but that is probably an artifact, due to the highly skewed distribution of importance values.

Please note that the experts agreed on the importance (or less importance) of 92% of the comments (158 from total of 172 comments), a number that reflects a high level of agreement, in contrast to the relatively low Kappa. In order to estimate the number of experts required for achieving higher Kappa of 0.7, we used the Spearman-Brown prediction formula, as suggested by [van Ast, 2004], and found that it will require to multiply the number of experts by 3.97, i.e., include eight experts.

The overall voting score for importance of the issues referred to by all of the comments was 93%, which is the portion of the number of judgments of a comment as important (320 votes) out of the number of all importance judgments (344 votes).

### 3.4. A Comparison of the Correctness and Completeness of the Experts and the System

As an additional aspect in the correctness and completeness analysis, we were interested in comparing the results of the different experts, and to compare their results to those of the system (See issue 1 in Section 2.1). Although the experiment did not include a step in which the experts directly evaluated the comments of each other, we could use their evaluations from the two experimental steps to indirectly calculate their level of completeness and correctness. For this, we have defined two additional objective measures to evaluate the quality of the experts' evaluations: Indirect Correctness and Indirect Completeness.

#### 3.4.1. Indirect correctness of the experts

The *Indirect Correctness* of an expert was calculated by using the comments that were mentioned by the expert in the first step of manual evaluation of the treatment's compliance, in which each expert added their comments regarding the compliance to the guideline as manifested in the medical records of each of the 10 patients. The indirect correctness is measured by the portion of the expert's comments that were mentioned by at least one additional expert.

It is important to note that the level of completeness of an expert has an effect on the results of the indirect correctness of the other experts, since an expert who makes only a relatively small number of the relevant comments artificially reduces the correctness of the comments made by the other experts. Due to this fact, we were interested in adding the system as an additional compliance evaluation agent. In the previous steps of the analysis, the completeness and correctness of the system were found relatively high; therefore, we assumed it is reasonable to use the comments of the system as part of the evaluations of the indirect correctness of the experts.

Table 4 displays the results of the indirect correctness analysis, when using the expert comment only, and when taking the system comments into consideration. For example, 99% of the comments mentioned by diabetes expert 1 were mentioned by at least one other agent (incl. the system), but only 86% were mentioned by one other expert (excl. the system). Notice that the indirect correctness results of all experts were higher when the system was added as an additional agent, due to the high completeness of the system.

**Table 4. Indirect correctness of the experts' comments, partitioned by level of support of the comments by the other agents, including the system.**

|  | Diabetes Expert 1 | Diabetes Expert 2 | Family Medicine Expert |
|---|---|---|---|
| All Comments | 86 | 118 | 125 |
| Not mentioned by any other agent / expert | **2** / 12 | **11** / 35 | **15** / 36 |
| Mentioned by 1 other agent / expert | **14** / 23 | **30** / 36 | **27** / 37 |
| Mentioned by 2 other agents / expert | **20** / 51 | **31** / 47 | **32** / 52 |
| Mentioned by 3 other agents / expert | **50** / NA | **46** / NA | **51** / NA |
| % comments mentioned by at least 1 other agent | **99%** / 86% | **91%** / 70% | 88% / 71% |

### 3.4.2. Indirect completeness of the experts

The *Indirect Completeness* of an expert was calculated using the evaluations from the second part of the experiment, in which the two diabetes experts separately evaluated the correctness of the compliance comments provided by the system, and thus, indirectly, assessed the comments given by the other experts. In this part of the evaluation, the expert judged 156 system comments as "jointly correct" (i.e., judged as correct by both experts or as correct by one and as partially correct by the other). The indirect completeness of an expert is measured by the portion of the "jointly-correct" comments that were mentioned by the expert in his manual compliance evaluation of the same patient. Table 5 displays the results of the indirect completeness analysis.

**Table 5. Indirect Completeness of the experts in the manual compliance evaluation.**

|  | Diabetes Expert 1 | Diabetes Expert 2 | Family Medicine Expert | All Experts |
|---|---|---|---|---|
| Judged as jointly correct and mentioned by the expert in his comments | 117 | 93 | 86 | 296 |
| Judged as jointly correct but not mentioned by the expert in his comments | 39 | 63 | 70 | 172 |
| Total "jointly correct" comments | 156 | 156 | 156 | 468 |
| Indirect Completeness relative to the "jointly correct" comments | 75% | 60% | 55% | 63% |

### 3.4.3. Comparison between the experts and the system

To conclude the completeness and correctness analysis, we performed a comparison between the results of all experts and the results of the system.

Table 6 and Figure 4 summarize the completeness and correctness results for the system and all experts. For the system we used the results mentioned in the completeness and correctness sections of the analysis, 91% for completeness and 91% for correctness. For the experts we used the results of the indirect completeness and indirect correctness presented in the previous sections. Diabetes expert 1 had the highest correctness score of 99%, and the system had the highest completeness score of 91% with a correctness score of 91%, which is similar to the correctness score of diabetes expert 2. When using the Harmonic Mean to integrate the correctness and completeness, the system resulted with the highest score of 0.91. The Harmonic Mean is defined as:

$$Harmonic\ Mean = 2 * (Correctness * Completeness) / (Correctness + Completeness)$$

**Table 6.** Summary of completeness and correctness of the system and the experts.

|  | Completeness (%) | Correctness (%) | Harmonic Mean |
|---|---|---|---|
| System | **91** | 91 | **0.91** |
| Diabetes Expert 1 | 75 | **99** | 0.85 |
| Diabetes Expert 2 | 60 | 91 | 0.72 |
| Family Medicine Expert | 55 | 88 | 0.68 |

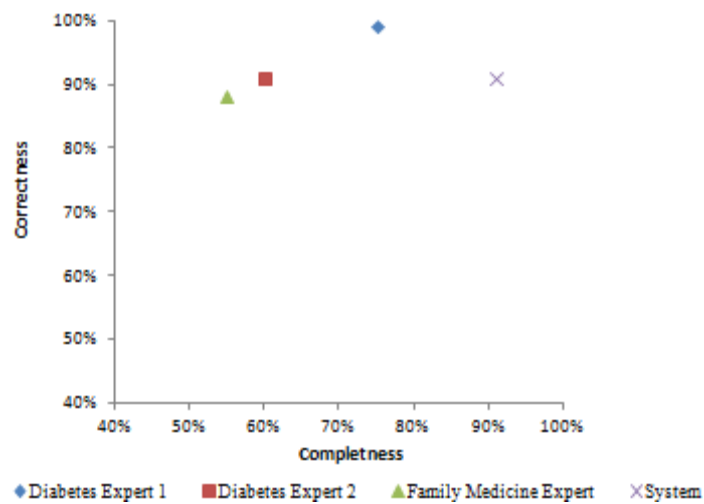

**Figure 4.** A profile of the completeness and correctness of the experts and the system.

## 4.   Summary and Discussion

The results of the evaluation provided answers to all of the three major research questions of this research, as well as to the two secondary issues raised in Section 2.1:

The *DiscovErr* system was found to produce most (91%) of the important compliance-related comments, when applied to real patients; its completeness score was actually higher than all medical experts participating in the experiment.

Most of the comments given by the system were found correct when assessed by the medical experts, and the system achieved a correctness score (91%) which was similar to the score of one of the diabetes experts; it was in fact higher than the score of the family medicine expert, although lower than the score of the second diabetes expert.

The compliance-related comments of the system were found significant and important, when directly evaluated by the diabetes experts regarding this aspect.

With respect to the secondary issues examined during the evaluation, 46% of the unique comments regarding compliance issues (with respect to content) were made by only one expert; 26.5% were made by two; and the rest, 27.5%, were made by all three experts. The inter-agreement between the diabetes experts regarding the correctness of the system comments, assessed through a weighted Kappa measure, is considered good and significant 0.61 ($p < 0.05$).

It is interesting to note, that beyond the significant weighted-Kappa value of the meta-agreement amongst the experts, i.e., regarding the correctness of the system comments, the diabetes experts fully or partially agreed on the correctness, partial correctness, or incorrectness of 171 of the 172 evaluated system comments (99.4%).  This high level of inter-observer agreement regarding the meta-critiquing task was rather surprising to us, but we found it quite encouraging, with respect to supporting the evaluation – it would have been more difficult to assess the correctness of the comments made by the *DiscovErr* system if the agreement among the experts had been rather low. Nevertheless, note that the high agreement amongst the two diabetes experts was regarding the results of the meta-critiquing task, namely, assessing the critiquing comments of an external agent, i.e., the *DiscovErr* system, regarding a given therapy by the care provider of the patient; we did not measure agreement regarding the recommended optimal therapy itself.

The fact that the performance of the *DiscovErr* system had a higher completeness when compared to the medical experts, can be explained by the advantage of the computer in performing such tasks, which involve scanning a relatively large amount of temporal data in order to analyze the compliance to the guideline. As previously mentioned in the results, the experts required 27 minutes, on average, to evaluate the medical record of a single patient, and the medical records contained an average number of 158 time-stamped data items, describing about five years of medical treatment. In these types of tasks, the computer has a clear advantage on human experts, who may miss certain compliance problems. It is important to mention that the experts stated that the user interface provided for them, which visualizes the temporal data of multiple parameters in parallel graphs, was very useful, and that if they were required to perform this task using the systems they currently use in the real clinical settings, it would have been much harder, almost impossible.

It is encouraging to note that the *DiscovErr* system produced 98% of the comments made by all three experts (versus 83% of the comments made by only two experts, and 66% of the comments made by only one expert). In our opinion, this result provides, in addition to the explicit assessment of the system comments by the experts, yet another implicit validation that the system focuses on important issues.

We have a reasonable expectation that the *DiscovErr* system might perform at a similar level in other time-oriented clinical domains, such as in the management of other types of chronic patients, monitoring of pregnancies, and even the management of patients in an intensive-care unit. This expectation, although not assessed in the current study, is based on the fact that these domains share similar characteristics of data and knowledge, and on the fact that the underlying knowledge-representation model that the *DiscovErr* system uses is based on the Asbru procedural specification language and on the KBTA declarative-knowledge ontology. The expressiveness of these representation formats has been assessed by multiple studies in the past in various clinical domains. For example, the Asbru language has been used by other systems for critiquing and for quality assessment [Advani et al., 2002; Sips et al., 2006; Boldo 2007; Groot et al., 2008] and its capability for formal representation of clinical guidelines was assessed in several projects as described in the introduction [Fuchsberger et al., 2005; Shalom et al., 2015; Peleg et al., 2017]. Furthermore, as there is no domain-specific element in the compliance analysis algorithm we had presented, it is reasonable to assume that it will work well in other clinical domains. Nevertheless, this assertion needs to be verified in an additional future research.

**Implications to Quality Assessment of Medical Care**

Note that the compliance analysis performed retrospectively in this study by the *DiscovErr* system can also be performed in real time, at the point of care, by assessing the quality of the healthcare provider decisions, as opposed to actions, to immediately assist clinicians in increasing their compliance when deviations from the guidelines are detected. Such a critiquing mode, "over the shoulder" style of guideline-based support aims to provide on-line decision support with minimal interaction with the clinician, thus enhancing the acceptance of decision-support systems in real clinical settings.

In recent years, healthcare providers have invested increasing efforts in applying methods and measures to assess the quality of the medical care that they provide for their patients. Examples of such quality measures are the Clinical Quality Measures (CQMs) published by the Centers for Medicare & Medicaid Services (CMS), and the Indicators for Quality Improvement (IQIs) published by the NHS. Efforts are also being made for the development of automated systems for reporting the compliance to these measures, including the development of quality data models by organizations such as the National Quality Forum (NQF), for the linking of local databases to the standards of the publishers [Dykes et al., 2010], and for the development of tools to improve the quality of data in order to support automated analysis of these measures [Lanzola et al., 2014]. These types of measures can be easily represented using the knowledge model of *DiscovErr*, and then can be automatically applied to medical records by the system. However, the methodology for automated compliance analysis is designed to represent and assess much more complex measures, which consider also the temporal patterns formed by the longitudinal therapy and not just specific points in the life of the patient (such as what happened on a particular clinic visit). For example, one might wish to consider the pattern formed by all of the LDL-C measurements since the start of therapy, and verify that it is indeed gradually decreasing, and not consider only the most recent values).

**Limitations of the Results**

Due to limited resources and time, the evaluation in this study was performed in the single medical domain of type 2 diabetes, although, as mentioned above, the *DiscovErr* system is generic and might be used for analysis of compliance to multiple clinical guidelines in multiple medical domains. In addition, adding more medical experts and extending their evaluation to include medical records of additional patients, could have increased the statistical significance of the results, if such experts and their time were available to us.

# 5.   Conclusions

We conclude that systems such as *DiscovErr* can be effectively used to provide expert-level critique of longitudinal evidence-based care, and to effectively assess the quality of time-oriented guideline-based care of large numbers of patients.

# Acknowledgements

Dr. Hatsek's research was partially supported by the Israeli National Institute for Health Policy Research.